\pdfoutput=1

\documentclass[11pt]{article}

\usepackage{ACL2023}
\usepackage{hyperref}
\usepackage{times}
\usepackage{latexsym,booktabs}
\usepackage{todonotes}

\usepackage{multirow,amsfonts}

\usepackage[T1]{fontenc}

\usepackage[utf8]{inputenc}

\usepackage{microtype}
\usepackage{amsmath}
\usepackage{inconsolata}

\title{Mapping Brains with Language Models: A Survey}

\newcommand{\ku}{$^1$}
\newcommand{\pr}{$^2$}

\author{Antonia Karamolegkou\ku, Mostafa Abdou\pr, Anders S{\o}gaard\ku \\
{\ku}University of Copenhagen,
{\pr} Princeton University\\ \texttt{antka@di.ku.dk}, \texttt{ma4231@princeton.edu}, \texttt{soegaard@di.ku.dk}
}

\begin{document}
\maketitle
\begin{abstract}

Over the years, many researchers have seemingly made the same observation: Brain and language model activations exhibit {\em some} structural similarities, enabling linear partial mappings between features extracted from neural recordings and computational language models. In an attempt to evaluate how much evidence has been accumulated for this observation, we survey over 30 studies spanning 10 datasets and 8 metrics. How much evidence has been accumulated, and what, if anything, is missing before we can draw conclusions? Our analysis of the evaluation methods used in the literature reveals that some of the metrics are less conservative. We also find that the accumulated evidence, for now, remains ambiguous, but correlations with model size and quality provide grounds for cautious optimism.

\end{abstract}

\section{Introduction}

Advances in neuroimaging technologies have made it possible to better approximate the spatiotemporal profile of the computations responsible for language in the brain \cite{Poldrack2015, neoroimaging}. At the same time, advances in natural language processing have produced language models (LMs) with high performance in many tasks \citep{nlp_survey}.

This progress has motivated scientists to start using state-of-the-art LMs to study neural activity in the human brain during language processing \citep{wehbe-etal-2014-aligning,Huth2016-ww,Schrimpfdoi:10.1073, Toneva_brain_cause2022,Caucheteuxpartialconverge}. Conversely, it has also prompted NLP researchers to start using neuroimaging data to evaluate and improve their models \citep{sogaard-2016-evaluating,bingel-etal-2016-extracting,Toneva2019InterpretingAI,hollenstein-etal-2019-cognival,loong_NLU_brain}.

At the conceptual core of these studies lies the suggestion that representations extracted from NLP models can (partially) explain the signal found in neural data. %
These representations can be based on co-occurrence counts \citep{mitchell_2008,Pereira2013UsingWT,Huth2016-ww} or syntactic and discourse features \citep{Wehbe2014-simultaneously,wehbe-etal-2014-aligning}.  Later studies use dense representations such as word embeddings \citep{anderson-etal-2017-visually,pereira2018,Toneva2019InterpretingAI,hollenstein-etal-2019-cognival} %
and recurrent neural networks to extract contextual stimuli representations \citep{Qian2016BridgingLA,Jain_Huth_2018_context,sun_wang_2019_decoding}. More recently, transformer-based architectures have been shown to align even better with neural activity data \cite{gauthier2019linking, Schrimpfdoi:10.1073, oota-etal-2022-tasks}. 

Such work shows that LMs can be trained to induce representations that are seemingly predictive of neural recordings or features thereof. However, pursuing the literature, it quickly becomes clear that these papers all rely on different experimental protocols and different metrics \citep{minnema-herbelot-2019-brain, hollenstein-etal-2020, Beinborn2019RobustEO}. So questions are: 
How much evidence has really been accumulated in support of structural similarities between brains and LMs? And more importantly, what exactly, if anything, drives this alignment, and what are we to understand from it? After gathering all the studies, we examine their evaluation metrics and their interrelationships, providing discussions on the corresponding findings.

\paragraph{Contributions} Our study provides four major contributions for the wider NLP audience: (a) a detailed review of the literature on mappings between fMRI/MEG recordings and representations from language models; 
(b) an overview of the datasets and mapping methods;
(c) an analysis of the evaluation setups that have been used to link neural signals with language models and how they relate; (d) a discussion of what drives this representational alignment and what we, as a field, can make of it going forward.

\paragraph{Terminology} First, a brief note on terminology: %
{Neural response measurements} refer to recordings of the brain activity of subjects reading or listening to language. We focus on  %
(a) functional magnetic resonance imaging (fMRI), which measures neuronal activity via blood oxygenation level-dependent contrast and has a high spatial resolution but poor temporal resolution (3--6s) and (b) magnetoencephalography (MEG), which involves the measurement of the magnetic field generated by the electrical activity of neurons in the cortex, providing a more accurate resolution of the timing of neuronal activity. %

{Voxels} refer to the smallest unit of data %
in a neuroimage, being the three-dimensional equivalent of a pixel in two-dimensional images \citep{Gerber2008-iq}. 
Finally, we use {brain decoding} to refer to predicting stimuli from brain responses (i.e. 
\textit{reading the brain}). %
{Brain encoding} will then refer to predicting brain responses from stimuli. Whereas decoding models serve as a test for the presence of information in neural responses, encoding models can be interpreted as process models constraining brain-computational theories %
\citep{decoding_encoding_Kriegeskorte}.

\section{Datasets}

To infer a mapping between language models and brains, researchers rely on datasets in which brain activity is recorded in response to linguistic stimuli. In some studies, the stimuli are single words \citep{mitchell_2008,anderson-etal-2017-visually}
or sentences displayed on a screen \citep{pereira2018}. In others, participants read longer stories \citep{Wehbe2014-simultaneously,alice,narratives} or listened to speech or podcasts \citep{Huth2016-ww,Antonello2021LowDimensionalSI}. Table~\ref{datasets_table} lists publicly available datasets that have been used in the context of mapping language models to and from recordings of brain response. Differences between the datasets --the number of participants, the equipment, the experimental setup, pre-processing steps, and probabilistic corrections -- should lead us to expect some variation in what researchers have concluded \citep{hollenstein-etal-2020}. %

\begin{table*}
\centering\small 
\begin{tabular}{llllll}
\hline
&\textbf{Data} & \textbf{Authors} & \textbf{Method}  & \textbf{N subjects}   \\
\hline

  1&\href{https://www.cs.cmu.edu/afs/cs/project/theo-73/www/science2008/data.html}{60 Nouns} & \citet{mitchell_2008} & fMRI  & 9\\
  
    2&\href{http://www.cs.cmu.edu/~fmri/plosone/}{Harry Potter Dataset} & \citet{Wehbe2014-simultaneously} & fMRI  & 8\\

        3&\href{http://www.cs.cmu.edu/~fmri/plosone/}{Harry Potter Dataset} & \citet{wehbe-etal-2014-aligning} & MEG  & 8\\

    4&\href{https://gallantlab.org/huth2016/}{The Moth Radio Hour Dataset} & \citet{Huth2016-ww} & fMRI  & 7\\

      5&\href{https://osf.io/crwz7/}{Pereira Dataset} & \citet{pereira2018} & fMRI & 16\\
  6&\href{https://data.donders.ru.nl/collections/di/dccn/DSC_3011020.09_236?0}{Mother of all Unification
Studies} & \citet{dutchmother} & fMRI, MEG  & 204\\

7&\href{https://osf.io/eq2ba/}{Natural Stories Audio Dataset} & \citet{Zhang649939} & fMRI & 19\\
 
  8&\href{http://datasets.datalad.org/?dir=/labs/hasson/narratives}{Narratives} & \citet{narratives} & fMRI & 345  \\

   9&\href{https://openneuro.org/datasets/ds003020/versions/1.0.2}{Podcast dataset} & \citet{Antonello2021LowDimensionalSI} & fMRI & 5  \\

 10&\href{https://openneuro.org/datasets/ds003643/versions/2.0.1}{The Little Prince Datasets} & \citet{LittlePrince} & fMRI & 112 \\

\hline
\end{tabular}
\caption{\label{datasets_table} The opensource datasets that were used in the studies surveyed. %
Numbering used in Table~2.}
\end{table*}

\section{How to predict brain activity?}
In this section, we survey work in which neural responses are predicted from linguistic representations. Such work typically aims to shed light on how language functions in the brain. %
One of the earliest studies exploring the mapping between brain and language representations is by \citet{mitchell_2008}, who trained a linear regression model on a set of word representations extracted from 60 nouns using 115 semantic features based on co-occurrence statistics, to predict the corresponding fMRI representations of the same nouns. They use pair-wise matching accuracy evaluation, extracting two words $w$ and $w'$ for evaluation, and showed that the predicted fMRI for a word $w$ was closer to the real fMRI image for $w$ than to the real fMRI image for $w'$, at above-chance levels. %
\citet{mitchell_2008} also report percentile rank results, ranking predicted fMRI images by similarity with the real image of $w$. We discuss how the metrics relate in \S6.  %

The dataset of \citet{mitchell_2008}is also used by \citet{murphy-etal-2012-selecting}, who extract linguistic features from  part-of-speech taggers, stemmers, and dependency parsers, showing that dependency parsers are the most successful in predicting brain activity. They also use leave-2-out pair-matching as their performance metric. %

Later on, \citet{Wehbe2014-simultaneously} moved on to predicting brain activation patterns for entire sentences rather than for isolated words. They recorded fMRI neural response measurements while participants read a chapter from {\em Harry Potter and the Sorcerer’s Stone}, then extracted a set of 195 features for each word (ranging from semantic, syntactic properties to visual and discourse-level features) to train a comprehensive generative model that would then predict the time series of the fMRI activity observed when the participants read that passage. Leave-2-out pair-matching accuracy is used for evaluation. %

\citet{Huth2016-ww}, in contrast, use fMRI recordings of participants listening to spoken narrative stories, representing each word in the corpus as a 985-dimensional vector encoding semantic information driven by co-occurrence statistics. They train per-voxel linear regression models and evaluate their predicted per-word fMRI images by their per-voxel Pearson correlation with the real fMRI images, showing that 3-4 dimensions explained a significant amount of variance in the FMRI data. %

\citet{wehbe-etal-2014-aligning} are among the first to use neural language models, using {\em recurrent} models to compute contextualized embeddings, hidden state vectors of previous words, and word probabilities. They run their experiments of MEG recordings of participants reading Harry Potter, obtained in a follow-up study to \citet{Wehbe2014-simultaneously}. From the three sets of representations, they then train linear regression models to predict the MEG vectors corresponding to each word, and the regression models are then evaluated by computing pair-matching accuracy. %

Similarly, \citet{sogaard-2016-evaluating} evaluates static word embeddings on the data from \citet{Wehbe2014-simultaneously}, learning linear transformation from word embeddings into an fMRI vector space. The predictions are evaluated through mean squared error (MSE). 

\citet{Jain_Huth_2018_context} evaluate recurrent language models %
against the fMRI dataset from \citet{Huth2016-ww}. Their findings show that contextual language model representations align significantly better (to brain response) compared to static word embedding models. Their evaluation metric is the total sum of explained variance\footnote{The squared Pearson correlation coefficient. We will not distinguish between studies using Pearson correlation and studies using explained variance. See Appendix \ref{sec:appendix2}.}

Following this, \citet{Schwartz2019InducingBB} use attention-based transformer language models for brain mapping. They finetune BERT \cite{devlin-etal-2019-bert}to predict neural response measurements from the Harry Potter dataset, showing that the fine-tuned models have representations that encode more brain-activity-relevant language information than the non-finetuned models. They rely on pair-matching accuracy as their performance metric. %

As in \citet{sogaard-2016-evaluating},  \citet{Zhang649939} map static word embeddings into the vector space of the neural response measurements (fMRI). They introduce a new dataset of such measurements from subjects listening to natural stories. %
They rely on explained variance as their performance metric. %

\citet{Toneva2019InterpretingAI} evaluate word and sequence embeddings from 4 recurrent and attention-based transformer language models, using the Harry Potter fMRI dataset. They evaluate models across layers, context lengths, and attention types, using pairwise matching accuracy as their performance metric. In a later study, \citet{Toneva2022combining} induce compositional semantic representations of "supra-word meaning" which they then use to predict neural responses across regions of interest, evaluating their models using Pearson correlation. 

Also using the Harry Potter data, \citet{abnar-etal-2019-blackbox} evaluate five models, one static and four contextualized, relying on a variant of representational similarity analysis \citep{Kriegeskorte2008-RSA}. %
The results suggest that models provide representations of local contexts that are well-aligned to neural measurements. However, as information from further away context is integrated by the models, representations become less aligned to neural measurements.

In a large-scale study, \citet{Schrimpfdoi:10.1073} examine the relationships between 43 diverse state-of-the-art neural network models (including embedding models, recurrent models, and transformers) across three %
datasets (two fMRI, one electrocardiography). They rely on a metric they term Brain Score which involves normalising the Pearson correlation by a noise ceiling. Their results show that %
transformer-based models perform better than recurrent or static models, and larger models perform better than smaller ones. %

Similarly, in \citet{Caucheteuxpartialconverge}, the \citet{dutchmother} fMRI and MEG datasets are used to compare a variety of transformer architectures. They study how architectural details, training settings, and the linguistic performance of these models independently account for the generation of brain correspondent representations. The results suggest that the better language models are at predicting words from context, the better their activations linearly map onto those of the brain. 

\citet{Antonello2021LowDimensionalSI} evaluate three static and five attention-based transformer models, in combination with four fine-tuning tasks and two machine translation models. %
They train linear regression models to evaluate their word-level representations %
against a new fMRI dataset from participants listening to podcast stories. They find a low-dimensional structure in language representations that can predict brain responses. In a similar setting, \citet{antonello_why} examine why some features fit the brain data better arguing that the reason is that they capture various linguistic phenomena. %

\citet{Reddy_syntax} evaluate syntactic features in conjunction with BERT representations, finding that syntax  explains additional variance in brain activity in various parts of the language system, even while controlling for complexity metrics that capture processing load. %

In a series of studies \citet{Caucheteux2021DisentanglingSA, Caucheteux_semantic, Caucheteux22LongrangeAH} investigate GPT2's activations in predicting brain signals using the \citet{narratives} dataset. %
Their evaluation metric is Brain Score \cite{Schrimpf_brain_score_2018}.
To determine which factors affect the brain encoding \citet{pasquiou} examine the impact of test loss, training corpus, model architecture, and fine-tuning in various models using the \citet{LittlePrince} dataset. They evaluate model performance using Pearson Correlation.

\citet{oota2022_encod} study the impact of context size in language models on how they align with neural response measurements. They use the \citet{narratives} dataset and evaluate recurrent and attention-based transformer architectures. In a later study, \citet{oota-etal-2022-tasks} use the \citet{pereira2018} dataset and evaluate BERT-base models (fine-tuned for various NLP tasks). %
They showed that neural response predictions from ridge regression with BERT-base models fine-tuned for coreference resolution, NER, and shallow syntactic parsing explained more variance for \citet{pereira2018} response measurements. %
On the other hand, tasks such as paraphrase generation, summarization, and natural language inference led to better encoding performance for the \citet{narratives} data (audio). %
Using the same dataset, in \citet{oota2022ling} it is shown that the presence of surface, syntactic, and semantic linguistic information is crucial for the alignment across all layers of the language model. 
They use pairwise matching accuracy and/or Pearson correlation as their performance metrics in these studies. %

\citet{loong_NLU_brain} extract feature representations from four attention-based transformer models. They evaluate the impact of fine-tuning on the BookSum dataset \citep{Kryscinski2021BookSumAC}. All models are used to predict brain activity on the Harry Potter data. Pairwise matching accuracy and Pearson correlation are their performance metrics. %
\citet{merlin_2022} focus more narrowly on variants of GPT-2, showing that improvements in alignment with brain recordings are probably not because of the next-word prediction task or word-level semantics,  but due to multi-word semantics. Their reported metric is Pearson correlation. %

\paragraph{Intermediate summary} The above studies differ in many respects. Several metrics are used: pairwise-matching accuracy,\footnote{Some papers \cite{wehbe-etal-2014-aligning,Toneva2019InterpretingAI,loong_NLU_brain} use a variant of pairwise-matching accuracy, in which the model has to discriminate between two averages of 20 random predicted neural response measurements. We do not distinguish between the two variants.} Pearson correlation (or Brain Score), mean squared error, and representational similarity analysis. Even studies that report the same performance metrics are not directly comparable because they often report on results on different datasets and use slightly different protocols, e.g., \citet{murphy-etal-2012-selecting} and \citet{wehbe-etal-2014-aligning}. \citet{Beinborn2019RobustEO} compare various encoding experiments and receive very diverse results for different evaluation metrics. The diversity of metrics and data renders a direct comparison difficult. To remedy this, we consider how the metrics compare in \S6. %

\section{How to predict linguistic stimuli?}
Decoding models work in the other direction and aim to predict linguistic features of the stimuli from recordings of brain response. 
\citet{pereira2018} introduce a decoder that predicts stimuli representation of semantic features given fMRI data.%
They introduce a novel dataset of neural responses aligned with annotation of %
concrete and abstract %
semantic categories (such as pleasure, ignorance, cooking etc.). %
They evaluate static word embeddings by applying ridge regression to predict per-word fMRI vectors. A separate regression model is trained per dimension, allowing for dimension-wise regularization. The model is evaluated in terms of pairwise matching accuracy, but also in terms of percentile rank, adapted to the decoding scenario. %

\citet{gauthier2019linking} 
also train linear regression models which map from the response measurements in \citet{pereira2018}, but to representations of the same sentences produced by the BERT language model finetuned on different natural language understanding tasks. The regression models are evaluated using two metrics: mean squared error and average percentile rank. %
Their results show that fine-tuning with different NLU objectives leads to worse alignment and that, somewhat surprisingly, the only objective which does lead to better alignment is a scrambled language modeling task where the model is trained to predict scrambled sentences.  

\citet{minnema-herbelot-2019-brain} re-examine the work of \citet{pereira2018} using various metrics (pairwise matching accuracy, percentile rank, cosine distance, $R^2$, RSA), comparing decoder models (ridge regression, perceptron, and convolutional neural networks).\footnote{Only the former two are linear and relevant for this meta-study.} They show that positive results {\em are only obtained using pairwise matching accuracy}. %

\citet{Abdou2021DoesIL} investigate whether aligning language models with brain recordings can be improved by biasing their attention with annotations from syntactic or semantic formalisms. They fine-tune the BERT models using several syntacto-semantic formalisms and evaluate their alignment with brain activity measurements from the \citet{Wehbe2014-simultaneously} and \citet{pereira2018} datasets. Their results -- obtained using Pearson correlation as performance metric -- are positive for two in three formalisms. %

\citet{zou-etal-2022-cross} propose a new evaluation method for decoding, a so-called {\em cross-modal cloze task}. They generate the data for the task from the neural response measures in \citet{mitchell_2008} and \citet{Wehbe2014-simultaneously}. The task itself amounts to a cloze task in which the context is prefixed by the fMRI image of the masked word. They evaluate models using precision@$k$. Note how this task is considerably easier than linearly mapping from language model representations into fMRI images, and precision@$k$ results therefore cannot be compared to those obtained in other settings. Their best precision@1 scores are around 0.3, but only marginally (0.03) better than a unimodal LM. %

Finally, \citet{Pascual2022} try a more realistic setup by predicting language from fMRI scans of subjects not included in the training. They use the \citep{pereira2018} dataset and evaluate the regression models based on pairwise accuracy and precision@k (or top-k accuracy). They propose evaluating with direct classification as a more demanding setup to evaluate and understand current brain decoding models.

\paragraph{Intermediate summary} Decoding studies also differ in many respects. Several metrics are used: pairwise-matching accuracy, Pearson correlation, percentile rank, cosine distance, precision@$k$, and representational similarity analysis; and several datasets are used. \citet{Gauthier2018} criticize the evaluation techniques of decoding studies and suggest adopting task and mechanism explicit models. It is of particular interest to our study that both \citet{minnema-herbelot-2019-brain} only report positive results for pairwise matching accuracy compared to other metrics. This suggests pairwise matching accuracy is a less conservative metric (and maybe less reliable). %

\begin{table*}
\centering\small 
\begin{tabular}{p{1.9in}p{0.4in}ccccccccc}
       \toprule {\textbf{Authors}} &{\bf Data}&  
       \textbf{E/D} &
       \textbf{Acc}  & 
       \textbf{P/B} & 
       \textbf{Rank} &
       \textbf{MSE} &
       \textbf{RSA} &
       \textbf{Cos Sim} & {\bf P@$k$}  \\ \midrule 
 \citet{mitchell_2008}  & 1 &E & \checkmark & & \checkmark  &  & &  \checkmark  \\ 
 \citet{murphy-etal-2012-selecting} & 1 &E & \checkmark &  \\
\citet{Wehbe2014-simultaneously, wehbe-etal-2014-aligning} & 2,3 &E  &  \checkmark &  \checkmark          \\
\citet{Huth2016-ww} &4 &E &   &  \checkmark\\
\citet{sogaard-2016-evaluating} & 2 & E & & & &\checkmark\\
\citet{pereira2018} & 5 &D & \checkmark & & \checkmark &\\
\citet{Jain_Huth_2018_context} & 4 &E & & \checkmark & & & & &   \\
\citet{Toneva2019InterpretingAI} & 2 &E & \checkmark &  & & & & &  \\
\citet{gauthier2019linking} &5 &D &  &  & \checkmark & \checkmark & \checkmark & &  \\
\citet{minnema-herbelot-2019-brain} & 5 &D &  \checkmark &  & \checkmark & & \checkmark & \checkmark &  \\
\citet{Schwartz2019InducingBB} & 3 &E & \checkmark &  &  &  \\
\citet{abnar-etal-2019-blackbox} & 2 &E & & & & & \checkmark \\
\citet{Zhang649939} & 7 &E & \checkmark & \checkmark &  &  &  &  & &   \\
\citet{Schrimpfdoi:10.1073} & 5 &E & \checkmark & \checkmark &  &  &  &  & &   \\
\citet{Abdou2021DoesIL} &2,5 &D & & \checkmark\\
\citet{Reddy_syntax} & 2 &E & &\checkmark & & & & &  \\
\citet{Antonello2021LowDimensionalSI} & 9 &E &  & \checkmark &  &  &  &  & &   \\
\citet{antonello_why} & 9 &E &  & \checkmark &  &  &  &  & &   \\
\citet{Caucheteux2021DisentanglingSA, Caucheteux_semantic, Caucheteux22LongrangeAH} & 8 &E &  & \checkmark &  &  &  &  & &   \\
\citet{Caucheteuxpartialconverge} & 6 &E &  & \checkmark &  &  &  &  & &   \\
\citet{zou-etal-2022-cross} & 1,2 &D &  &  &  &  &  &  & \checkmark &   \\
\citet{oota2022_encod, oota-etal-2022-tasks, oota2022ling} & 8 &E & \checkmark & \checkmark &  &  &  &  & &   \\
\citet{pasquiou} &10 &E & \checkmark &  &  &  &  &  & &   \\
\citet{Toneva2022combining} & 2 &E & & \checkmark & \\

\citet{merlin_2022} & 2 &E & & \checkmark &  &  &  &  & &   \\
\citet{Pascual2022} & 5 &D & \checkmark &  &  &  &  &  & \checkmark &   \\
\citet{loong_NLU_brain} & 2 &E & \checkmark & \checkmark & \\

\bottomrule
\end{tabular}
\caption{\label{metrics_eval} Overview of what studies rely on what data and what performance metrics. See Table~1 for dataset numbering. {\bf E/D:} Encoding or decoding model {\bf Acc:} Pairwise matching accuracy. {\bf P/B:} Pearson correlation or BrainScore. {\bf Rank:} percentile rank. {\bf MSE:} mean squared error. {\bf RSA:} representational similarity analysis. {\bf CosSim:} cosine similarity. {\bf P@$k$:} precision@$k$. \citet{Schrimpfdoi:10.1073} used a non-public fMRI dataset, too.  
}
\end{table*}

\section{Performance Metrics}

We present the evaluation metrics used in the above studies and discuss how they relate. %
See Table~\ref{metrics_eval} for a summary of metrics and corresponding studies.

\citet{mitchell_2008} introduce \textbf{pairwise matching accuracy}. Because of their small sample size, they use a %
leave-2-out cross-validation, which later work also adopted. The metric is a binary classification accuracy metric on a balanced dataset, so a random baseline converges toward 0.5. Many studies have relied on this metric, both in encoding and decoding (see Table~\ref{metrics_eval}).\footnote{The method is often referred to as 2v2 Accuracy. The variant that averages across 20 images, is then referred to as 20v20 Accuracy.} %

{\bf Pearson correlation}Pearson correlation is another widely used metric in the studies surveyed above, measuring the linear relationship between variables, and providing insight into the strength and direction of their association. \citet{Huth2016-ww}, compute Pearson correlation between predicted and actual brain responses using Gaussian random vectors to test statistical significance. Resulting $p$-values are corrected for multiple comparisons within each subject using false discovery rate (FDR) \citep{FDR_eval}. Others have used Bonferroni correction \citep{Huth2016-ww} or block-wise permutation test \citep{Adolf2014-cz} to evaluate the statistical significance of the correlation \citep{Zhang649939}. Some report $R^2$ (explained variance) instead of or in addition to correlation coefficients \cite{minnema-herbelot-2019-brain,Reddy_syntax}. %
Others have adopted a more elaborate extension of Pearson correlation, namely BrainScore \citep{Schrimpf_brain_score_2018}. BrainScore %
is estimated on held-out test data, calculating Pearson’s correlation between model predictions and neural recordings divided by the estimated ceiling and averaged across voxels and participants. 

\textbf{Percentile rank} was first used for encoding \citep{mitchell_2008}, but can also be used for decoding \citep{pereira2018, gauthier2019linking, minnema-herbelot-2019-brain}. In encoding, the predicted brain image for $w$ is ranked along the predicted images for a set of candidate words $w'$ by their similarity to the real (ground truth) image for $w$. The average rank is then reported. For decoding, they rank word vectors rather than neural response images. Note the similarity metric is unspecified, but typically cosine distance is used. %

\textbf{Mean squared error}, the average of the squared differences between word vectors and neural responses, was first used for encoding in \citet{sogaard-2016-evaluating} on a held-out test split. It was also used by \citet{gauthier2019linking}.

\textbf{Representational similarity analysis}
(RSA) was introduced in \citet{Kriegeskorte2008-RSA} as a non-parametric way to characterize structural alignment between the geometries of representations derived from disparate modalities. RSA
abstracts away from activity patterns themselves and
instead computes representational similarity
matrices (RSMs), which characterize the information carried by a given representation method
through global similarity structure. A rank correlation coefficient is computed between RSMs derived from the two spaces, providing a summary statistic indicative of the overall representational alignment between them. Being non-parametric, RSA circumvents many of the various methodological weaknesses (such as over fitting, etc.). \citet{gauthier2019linking},  \citet{minnema-herbelot-2019-brain}, and \citet{abnar-etal-2019-blackbox} apply (variations of) RSA to investigate the relations between different model components, and then to study the alignment of these components with brain response. %

\textbf{Cosine similarity} was used in \citet{mitchell_2008} to select between the candidate images in pairwise matching accuracy, as well as in percentile rank and RSA, but the raw cosine similarities between predicted and real images or embeddings can also be used as a metric. 
\citet{minnema-herbelot-2019-brain} use this metric to quantify how close the predicted word vectors are to the target. Finally, \citet{zou-etal-2022-cross} use {\bf precision@$k$}, a standard metric in other mapping problems, e.g., cross-lingual word embeddings \cite{4264a46fd9e846e4a704b2d13002e521}. 

\paragraph{Comparisons} Most metrics are used to evaluate both encoding and decoding models (pairwise matching accuracy, Pearson correlation, percentile rank, MSE, RSA, cosine distance). %
Results for two of the most widely used metrics -- pairwise matching accuracy\footnote{When discriminating averages over 20 images \cite{wehbe-etal-2014-aligning}, scores are naturally lower.} and percentile rank -- tend to be around 0.7--0.8 with generally better results for more recent architectures and larger LMs. To draw conclusions across studies relying on different metrics, we need to investigate which metrics are more conservative, and how different metrics relate. 

\paragraph{Pairwise matching accuracy vs.~Pearson correlation}

It seems that pairwise matching accuracy tends to increase monotonically with Pearson correlation. Consider three sets of distances over corresponding point sets, A, B, and C. If A and B are more strongly linearly correlated than A and C, under an optimal linear mapping $\Omega$ (minimizing point-wise squared error distance), $\mathbb{E}[(a-b\Omega)^2]>\mathbb{E}[(a-c\Omega)^2]$. Even in this conservative setting in our synthetic experiments in Appendix \ref{sec:appendix}, the correlation between matching accuracy and percentile rank was very high, \textasciitilde{0.9}. 

\paragraph{Pairwise matching accuracy vs.~percentile rank} Both metrics have random baseline scores of 0.5, and 
they will converge in the limit. If $a$ has a percentile rank of $p$ in a list $\mathcal{A}$, it will be higher than a random member of $\mathcal{A}$ $p$ percent of the time. In our experiments in Appendix \ref{sec:appendix}, the correlation converges toward 1.0, with values consistently higher than 0.8 for $N=100$. 

\paragraph{Pairwise matching accuracy vs.~precision@$k$} are also positively correlated. Perfect score in one entails perfect score in the other, but precision@$k$ can of course be very small for very high values of pairwise matching accuracy (especially if the set of candidate words is big). Conversely, we can have saturation for high values of $k$, because matching accuracies higher than $\frac{n-k}{n}$ will mean near-perfect precision@$k$ scores. In practice, precision@$k$ (for low values of $k$) will be much more conservative, however. The correlation coefficient for $N=100$ (see Appendix \ref{sec:appendix}) tends to lie around 0.7.

\paragraph{Relative strength} Pairwise Matching Accuracy is a relatively permissible performance metric. To see this, consider the scenario in which all target words can be divided into two equal-sized buckets based on word length (number of characters). Say the neural responses capture nothing but this binary distinction between long and short words, but do so perfectly. Moreover, our mapping method, e.g., linear regression, learns this from training data. Now, from this alone, the pairwise matching accuracy will converge toward $\mu=0.75$, since our model will do perfectly (1.0) on half of the data, and exhibit random performance (0.5) on the other half. If the neural responses tracked word length (and not just the distinction between short and long words), performance would be even better. In other words, Pairwise Matching Accuracy scores around 0.7-0.8 (observed in the studies above) may only reflect very shallow processing characteristics. The fact that \citet{minnema-herbelot-2019-brain} only observed good results with this metric, led them to adopt a rather critical stance, for good reasons. 

Other metrics are clearly more conservative. For a set of $n$ candidate words, a random mapping will induce a precision@1-score of $\frac{1}{n}$. While hubs may inflate scores for larger values, the metric is extremely conservative for small values of $k$. However, only \citet{zou-etal-2022-cross} use this metric, and they modify the experimental protocol substantially, making the task much easier by providing additional input to a non-linear model. The small improvement from adding neural response input is interesting, but could potentially be explained by shallow processing characteristics. 

They argue that analogy testing would provide a better evaluation protocol: 
\begin{quote}
    one would ideally use standard metrics such as semantic relatedness judgment tasks, analogy tasks, etc. [but] this is not possible due to the limited vocabulary sizes of the available brain datasets
\end{quote}

Such evaluation {\em is} possible on small scale, though, and increasingly larger fMRI datasets are becoming available (see above). \citet{Zhang649939} have identified analogical reasoning in fMRI brain activation spaces. The analogies are computed using vector offset and probe the systematicity of how semantic relations are encoded. If a model encodes the capital-of relation systematically, we can retrieve the capital of Germany by subtracting the fMRI vector for 'Paris' from the sum of our the fMRI vectors for Germany and France. This is the same kind of analogical reasoning found in language models \cite{mikolov2013linguistic}. \citet{Garneau2021} show that the more language models satisfy analogies, the more isomorphic they are. 

So far, it seems that, with the possible exception of \citet{Zhang649939}, there is little evidence for structural similarities, beyond what could be induced by shallow processing characteristics, but what about all the studies that report strong Pearson correlations? Per-voxel correlation coefficients are low on average, but across the above studies, typically only around 4-40\% of the voxels exhibit significant correlations \citep{Huth2016-ww, Caucheteuxpartialconverge}. Since these correlations have been replicated across different datasets, they are generally not disputed, but could still reflect rather shallow processing characteristics. 

On a more positive note, several studies show that larger (and better) language models align better with neural response measurements \cite{Schrimpfdoi:10.1073,Caucheteuxpartialconverge}. This suggests that language models in the future may align even better with such measurements, possibly reflecting properties of deep processing. Such correlations with model quality and size are positive, making the results reported above more credible. 

Generally, the conclusions we can draw from the above studies are somewhat vague. There are two reasons for this: (i) Past studies have relied on permissible (pairwise matching accuracy) and ambiguous (Pearson correlation) performance metrics; and (ii) past studies have relied on small-sized datasets. We believe that {\em this calls for a meta-analysis of the above studies}. To provide grounds for such a meta-analysis, we have in this section taken steps to compare the metrics used in these studies. We leave it for future work to explore various ways effect sizes can be computed across these studies.

\section{Discussion} %

Many studies, summarized above, aim to compare language model representations with neural response measurements using linear mapping models. Our main reason to focus on linear mapping models is that they quantify the degree of structural similarity (isomorphism). %
Overall, results suggest that structural similarities between language models and neural responses exist. Furthermore, there is good evidence that alignment has correlated positively with model quality and model size, suggesting a certain level of convergence as language models improve. %

\paragraph{What drives alignment?} %
Is alignment driven by deep processing characteristics or by shallow textual characteristics? %
Classical candidates for shallow ones would be word length, frequency, regularity, and part of speech. \citet{mitchell_2008}, for example, only controlled for part of speech. 
Some authors have presented results to suggest that alignments are driven by syntactic or semantic factors \cite{Abdou2021DoesIL, Reddy_syntax, Caucheteux2021DisentanglingSA, Zhang649939}, whereas others have claimed some similarities reflect semantic phenomena \cite{Huth2016-ww, Caucheteux2021DisentanglingSA}. Others suggest that alignments reflect deeper similarities between model objectives and predictive processing in human brains \citep{Schrimpf_brain_score_2018, Caucheteux22LongrangeAH, Goldstein2020.12.02.403477}, but see \citet{antonello_why} for a critical discussion of such work. 

Linguistically-transparent
models that allow for a principled decomposition of a model’s
components into smaller linguistically meaningful units and models that move  towards possible neurobiological implementations of neural computation are likely to be key for answering this question \citep{Hale_survey, Oever2022-ly}. Given the plethora of interpretability methods recently developed, however, we believe that even models which are not intrinsically interpretable can be
useful toward this goal.

\paragraph{Do some models align better?} Most studies observe that better and larger, contextual models align better with neural responses \citep{Jain_Huth_2018_context, Caucheteuxpartialconverge}. Other improvements include %
fine-tuning on specific tasks \citep{oota-etal-2022-tasks, loong_NLU_brain}. 
\citet{pasquiou} outline the impact of model training choices. %

\paragraph{What metrics?} The inconsistent use of performance metrics makes it hard to compare and interpret the results reported in the literature \citep{Beinborn2019RobustEO}. We have shown that some metrics are perhaps {\em too}~permissible to detect structural similarities between language models and neural responses. We have argued that precision@$k$ is more conservative than most other metrics. \citet{minnema-herbelot-2019-brain} have proposed using analogy scores. 
In the limit (given sufficient analogies), perfect analogical accuracy implies isomorphism \cite{Garneau2021}. So do perfect precision@1 and perfect RSA scores. We, therefore, propose giving priority to these performance metrics, not to conflate shallow processing characteristics with deeper, more semantic properties. 

\paragraph{Meta-analysis?} Proper meta-analysis is currently hindered by the use of different metrics, but we have taken steps to relate these. %

\section{Conclusions}

We surveyed work on linear mappings between neural response measurements and language model representations, with a focus on metrics. In particular, we surveyed a broad range of 30 studies spanning across 10 datasets and 8 metrics. By examining the metrics, and relating them to one another, we attempt to critically assess the accumulated evidence for structural similarity between neural responses and language model representations. We find that similarities with existing models are limited to moderate, and there is a possibility they might be explained by shallow processing characteristics since there is no standardised methodology for employing controls, but also that positive correlations with model quality and size suggest that language models may exhibit deeper similarities with neural responses in years to come.

\section*{Limitations}
This work focuses on a specific view of the whole neuro-computational modeling field. We exclude specific angles of research such as non-linear models \citep{ruan-etal-2016-exploring, Qian2016BridgingLA, bingel-etal-2016-extracting, anderson-etal-2017-visually, oota_2018} since we want to evaluate the accumulated evidence for structural similarity (isomorphism) between neural responses and language models. \cite{Ivanova_2022} mention several advantages of using linear mapping models, they are more interpretable and more biologically plausible. They also provide an insightful discussion on mapping model choice, emphasizing the importance of estimating models' complexity over categorizing them as purely linear or nonlinear. 

Another limitation is that we do not include speech models \citep{Vaidya2022SelfsupervisedMO, Defossezaudio, millet2022toward} that have been used to map brain representations mostly due to coherency and page-limit restrictions.  %
The survey is also limited to fMRI and MEG data instead of other modalities for two many reasons: (i) fMRI and MEG are used as a combination in many studies \citep{Caucheteuxpartialconverge, Schrimpfdoi:10.1073, Toneva2022combining}, 
and (ii) they offer high spatial resolution
and signal reliability (fMRI) and better temporal and spatial resolution (MEG), making them suitable for NLP \citep{hollenstein-etal-2020}. 
For a survey in encoding and decoding models in cognitive electrophysiology, see  \citet{survey_Holdgraf2017-jn}.

\section*{Ethics Statement}

The use of publicly available data in this survey ensures compliance with ethical guidelines while acknowledging the initial consent provided by the participants for data capture and sharing. Participants' consent is a crucial ethical consideration in the collection and sharing of fMRI and MEG data, and the preservation of legal and ethical rights should always be prioritized. By upholding ethical principles, researchers can responsibly contribute to the field of brain encoding and decoding, advancing our understanding of neural processes without compromising individual rights and privacy. Researchers should ensure secure storage, anonymization, and limited access to sensitive neuroimaging data, adhering to data protection regulations and guidelines.

Furthermore, it is essential to prioritize the dissemination of research findings in a responsible manner, with clear and accurate communication that respects the limits and uncertainties of scientific knowledge. Openness and transparency in reporting methods, results, and interpretations contribute to the overall integrity of the research field. Additionally, fostering a culture of collaboration, respect, and acknowledgment of the contributions of participants, colleagues, and the wider scientific community promotes ethical conduct and responsible research practices in brain encoding and decoding. By adhering to these ethical principles, researchers not only advance scientific knowledge but also build public trust, enhance the societal impact of their work, and ensure the long-term sustainability and progress of the field.

\section*{Acknowledgements}
This work is supported by the Novo Nordisk Foundation. Antonia Karamolegkou was supported by the Onassis Foundation - Scholarship ID: F ZP 017-2/2022-2023’.

\bibliography{custom}
\bibliographystyle{acl_natbib}

\appendix

\section{Appendix}
\label{sec:appendix}

\subsection{Metric Correlations}
\label{sec:appendix}
We used the following synthetic experiment to estimate the correlations between some of the most widely used performance metrics: 

\begin{itemize}
    \item[(i)] Generate $n$ random numbers and sort them to produce the list $\mathcal{A}$. 
    \item[(ii)] Sample $\frac{n}{10}$ items $\mathcal{B}$ of $\mathcal{A}$ at random.
    \item[(iii)] For $\epsilon\in\{\frac{1}{100},\ldots,\frac{100}{100}\}$, evaluate $\mu_{b\in\mathcal{B}}$ for $\langle b,\epsilon\cdot b\rangle$ for all metrics.  
\end{itemize}

In other words, for a noise level $\epsilon$, we evaluate predicted images or word vectors $\epsilon\cdot b$ against true images or word vectors $b$ relative to a set of target images/vectors of 99 candidate words. This experiment is easily repeated to estimate reliable coefficients.

\subsection{Correlation - Explained Variance}
\label{sec:appendix2}

In this study, we do not distinguish between studies using Pearson correlation and studies using explained variance.

Pearson Correlation can be defined as:

$$r = \frac{\sum_{i=1}^{n} (x_i - \overline{x})(y_i - \overline{y})}{\sqrt{\sum_{i=1}^{n} (x_i - \overline{x})^2}\sqrt{\sum_{i=1}^{n} (y_i - \overline{y})^2}}$$

where:

\begin{itemize}
    \item r is the Pearson correlation coefficient between variables X and Y
    \item x\_i and y\_i are individual data points for variables X and Y
    \item $\overline{x}$ and $\overline{y}$ are the means of variables X and Y
    \item n is the sample size.
\end{itemize}

The proportion of variance explained by the correlation is represented by $r^2$. The correlation coefficient ($r$) measures the strength and direction of the linear relationship, while the coefficient of determination ($R^2=r^2$) represents the proportion of the variance explained by the independent variable(s) in the dependent variable.

\end{document}